\documentclass[sigconf,balance=false]{acmart}

\usepackage{wrapfig}
\usepackage{makecell}
\usepackage{enumitem}
\usepackage{soul}
\usepackage{capt-of}
\usepackage{xcolor}
\usepackage{xspace}
\usepackage{colortbl}
\usepackage{amsmath,amsfonts,dsfont,pifont,bm,bbm,mathrsfs,mathtools,nicefrac}
\usepackage{booktabs,multirow,multicol,adjustbox,diagbox,threeparttable,textcomp}
\usepackage{colortbl}
\definecolor{colorTrd}{rgb}{0.95,0.95,0.65}
\definecolor{colorSnd}{rgb}{1, 0.85, 0.7}
\definecolor{colorFst}{rgb}{1, 0.7, 0.7}
\definecolor{green}{rgb}{0.45, 0.62, 0.31}
\definecolor{blue}{rgb}{0.42, 0.60, 0.82}
\definecolor{yellow}{rgb}{0.80, 0.64, 0.22}

\AtBeginDocument{%
  }

\copyrightyear{2024}
\acmYear{2024}
\setcopyright{acmlicensed}\acmConference[SIGGRAPH Conference Papers
'24]{Special Interest Group on Computer Graphics and Interactive Techniques
Conference Conference Papers '24}{July 27-August 1, 2024}{Denver, CO, USA}
\acmBooktitle{Special Interest Group on Computer Graphics and Interactive
Techniques Conference Conference Papers '24 (SIGGRAPH Conference Papers
'24), July 27-August 1, 2024, Denver, CO, USA}
\acmDOI{10.1145/3641519.3657441}
\acmISBN{979-8-4007-0525-0/24/07}

\begin{document}

\title{High-quality Surface Reconstruction using Gaussian Surfels}

\author{Pinxuan Dai}
\authornote{Joint first authors}
\email{daipinxuan@zju.edu.cn}
\affiliation{%
  \institution{State Key Lab of CAD\&CG, Zhejiang University}
  \country{China}
}

\author{Jiamin Xu}
\authornotemark[1]
\email{superxjm@yeah.net}
\affiliation{%
  \institution{Hangzhou Dianzi University}
  \country{China}
}

\author{Wenxiang Xie}
\email{zju_xwx@zju.edu.cn}
\affiliation{%
  \institution{State Key Lab of CAD\&CG, Zhejiang University}
  \country{China}
}

\author{Xinguo Liu}
\email{xgliu@cad.zju.edu.cn}
\affiliation{%
  \institution{State Key Lab of CAD\&CG, Zhejiang University}
  \country{China}
}

\author{Huamin Wang}
\email{wanghmin@gmail.com}
\affiliation{%
  \institution{Style3D Research}
  \country{United States of America}
}

\author{Weiwei Xu}
\authornote{Corresponding author}
\email{xww@cad.zju.edu.cn}
\affiliation{%
  \institution{State Key Lab of CAD\&CG, Zhejiang University}
  \country{China}
}

\newcommand{\weiwei}[1]{{\color{black} {} #1}}
\newcommand{\jiamin}[1]{{\color{black} {} #1}}
\newcommand{\pinxuan}[1]{{\color{black} {} #1}}

\begin{abstract}
We propose a novel point-based representation, Gaussian surfels, to combine the advantages of the flexible optimization procedure in 3D Gaussian points and the surface alignment property of surfels. This is achieved by directly setting the z-scale of 3D Gaussian points to 0, effectively flattening the original 3D ellipsoid into a 2D ellipse. Such a design provides clear guidance to the optimizer. By treating the local z-axis as the normal direction, it greatly improves optimization stability and surface alignment. While the derivatives to the local z-axis computed from the covariance matrix are zero in this setting, we design a self-supervised normal-depth consistency loss to remedy this issue. 
\jiamin{Monocular normal priors and foreground masks are incorporated to enhance the reconstruction quality, mitigating issues related to highlights and background.} We propose a volumetric cutting method to aggregate the information of Gaussian surfels so as to remove erroneous points in depth maps generated by alpha blending. Finally, we apply screened Poisson reconstruction method to the fused depth maps to extract the surface mesh. Experimental results show that our method demonstrates superior performance in surface reconstruction compared to state-of-the-art neural volume rendering and point-based rendering methods.
\end{abstract}


\ccsdesc[500]{Computing methodologies~Shape modeling}
\keywords{3D Surface Reconstruction, Gaussian Surfels, Depth-normal Consistency}
\begin{teaserfigure}
\includegraphics[width=\textwidth]{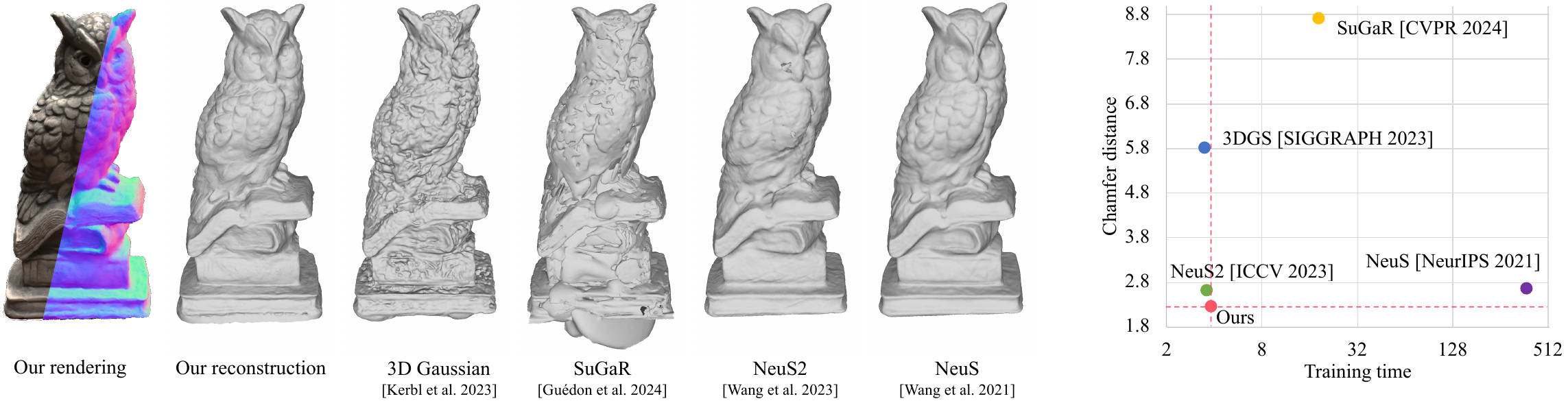}
\caption{Left: Our method achieves high-quality surface reconstruction result on par with NeuS\protect{~\cite{wang2021neus}} for this owl object (DTU-\#122). Right: The 2D plot shows that our method achieves a good balance between training time and the quality of surface reconstructions measured in the Chamfer distance. The statistics are collected from the scenes selected from BlendedMVS\protect{~\cite{yao2020blendedmvs}} dataset. }
\label{fig:teaser}
\end{teaserfigure}


\maketitle

\section{Introduction}


Recently, 3D Gaussian Splatting (3DGS) \cite{kerbl20233d} has gained widespread popularity for reconstructing and rendering 3D scenes. Different from neural implicit representations \cite{mildenhall2021nerf,wang2021neus, yariv2021volume}, 3DGS represents the appearance and geometry through a set of explicit and topology-free Gaussian points. These Gaussian points can be dynamically added and removed during optimization, ensuring the convergence of the optimization process to cover the entire surface and represent high-frequency details. Moreover, 3DGS utilizes GPU/CUDA-based rasterization with closed-form integration \cite{zwicker2002ewa} during rendering, eliminating the need for time-consuming ray-based point sampling in volume rendering. This results in a significant reduction in training and rendering time, enabling a training time of less than 10 minutes and facilitating real-time rendering.

Despite the advantages of the 3DGS representation, it struggles to generate high-quality geometric reconstructions. This limitation stems from three aspects: (1) \emph{Non-zero thickness}: 3D Gaussian points, resembling ellipsoids, have non-zero thickness along each axis, hindering their close alignment with the actual surfaces. (2) \emph{Ambiguity in normal direction}: there exists ambiguity in determining the normal for each 3D Gaussian, i.e. the normal axis can change among different scale directions during optimization. This ambiguity can result in inaccuracies when reconstructing geometries with fine details. (3) \emph{Modeling sharp surface edges}: the alpha blending process may introduce bias to a reconstructed surface edge, which will happen when Gaussian points with extent beyond the surface edge or far from the edge are occasionally involved during the blending. Recent methods such as SuGaR \cite{guedon2023sugar} and NeuSG \cite{chen2023neusg} introduce a regularizer to minimize the smallest component of the scaling factor along each axis, which can alleviate the first thickness problem. However, the quality of reconstructed surface is still unsatisfactory. Using Gaussian points to reconstruct surface with fine details remains challenging.

In this paper, we propose a novel representation, Gaussian surfels, to combine the advantages of the flexible optimization procedure in 3DGS and the surface alignment property of surfels \cite{pfister2000surfels}. This significantly improves the quality of the reconstructed geometry. Gaussian surfels are achieved by directly setting the z-scale of the scale matrix in 3D Gaussian points to 0, effectively flattening the original 3D ellipsoid into a 2D ellipse. Compared with the regularization methods in~\cite{guedon2023sugar,chen2023neusg}, our representation avoids the need to determine a minimal scale that might change during optimization. It provides clear guidance to the optimizer to treat the local z-axis as the normal direction, greatly improving optimization stability and surface alignment.

The technical challenge arising from flattening 3D Gaussian points in this manner is that the derivatives that are computed from the covariance matrix with respect to the local z-axis will be zero. As a result, the photometric loss itself can not affect the local z-axis during optimization. Therefore, we design a self-supervised normal-depth consistency loss to remedy this problem. It requires local z-axis to be close to the normal computed from the depth map rendered using Gaussian splatting. Moreover, although the truncation threshold in 3D Gaussian points is carefully selected to guarantee high-quality rendering results ($\frac{1}{255}$ in 3DGS), we found it is too small to prevent the generation of blurred sharp edges or floating geometries in the surface reconstruction results. These artifacts often occur when a ray passes through a front surface near its edge, but actually, the ray's first intersection with scene geometry is at a surface behind the front surface. We thus propose a volumetric cutting method to determine whether a voxel should be cut off or not according to its distance from the Gaussian surfels, which further improves geometric quality.



Our high-quality surface reconstruction results using Gaussian surfels are made possible by the following technical contributions:
\begin{itemize}[leftmargin=12pt,topsep=4pt]
\item Introducing a novel point-based representation, Gaussian surfels,  to resolve the inherent normal ambiguity of 3DGS and achieve a close alignment with the actual surface. Combined with our volumetric cutting method, the quality of surface reconstruction is significantly enhanced.

\item Proposing a self-supervised normal-depth consistency regularizer, along with the photometric loss, to guide the Gaussian surfels in moving and rotating in a manner that closely conforms to the surfaces of the object. 
We also incorporate monocular estimated normals as a prior to address shape-radiance ambiguity \cite{zhang2020nerf++} in regions with specular reflections.

\item Compared with state-of-the-art neural volume and point-based rendering methods, our method achieves a good balance between reconstruction quality and training speed (Fig. \ref{fig:teaser}).
\end{itemize}

\section{Related Work}

The target of multi-view surface reconstruction methods is to create a geometric representation of an object or a scene using multi-view images. This is accomplished through classic multi-view stereo (MVS) techniques \cite{furukawa2015multi}, which can be broadly classified as voxel grids optimization \cite{seitz1999photorealistic,sinha2007multi}, feature point growing \cite{furukawa2009accurate,wu2010quasi}, or depth-map estimation and merging \cite{schonberger2016pixelwise}. These methods rely on photometric consistency across views, making it challenging to accurately capture complete geometric representations due to the ambiguities in the correspondence.

Recently, neural rendering \cite{tewari2020state} has demonstrated impressive capabilities in view synthesis and surface reconstruction. By directly minimizing the per-pixel difference between the image and rendering results, it achieves detailed surface reconstruction \cite{wang2021neus}. Additionally, it can adapt to complex materials through a sophisticated rendering process \cite{zhang2021physg,zhang2022modeling}.

\subsection{Neural volume rendering}


The landmark work in neural volume rendering is NeRF~\cite{mildenhall2021nerf}, which employed a differentiable volumetric rendering technique to reconstruct a neural scene representation, achieving impressive photorealistic view synthesis with view-dependent effects. To accelerate its optimization, subsequent research replaces the neural scene representation with explicit or hybrid scene representations, such as voxel grid~\cite{sun2021direct, yu2021plenoxels}, low-rank tensors~\cite{chen2022tensorf}, tri-planes~\cite{chan2022efficient,reiser2023merf}, and multi-resolution hash maps \cite{muller2022instant}. 

\begin{figure*}[t]
\centering
\includegraphics[width=1\linewidth]{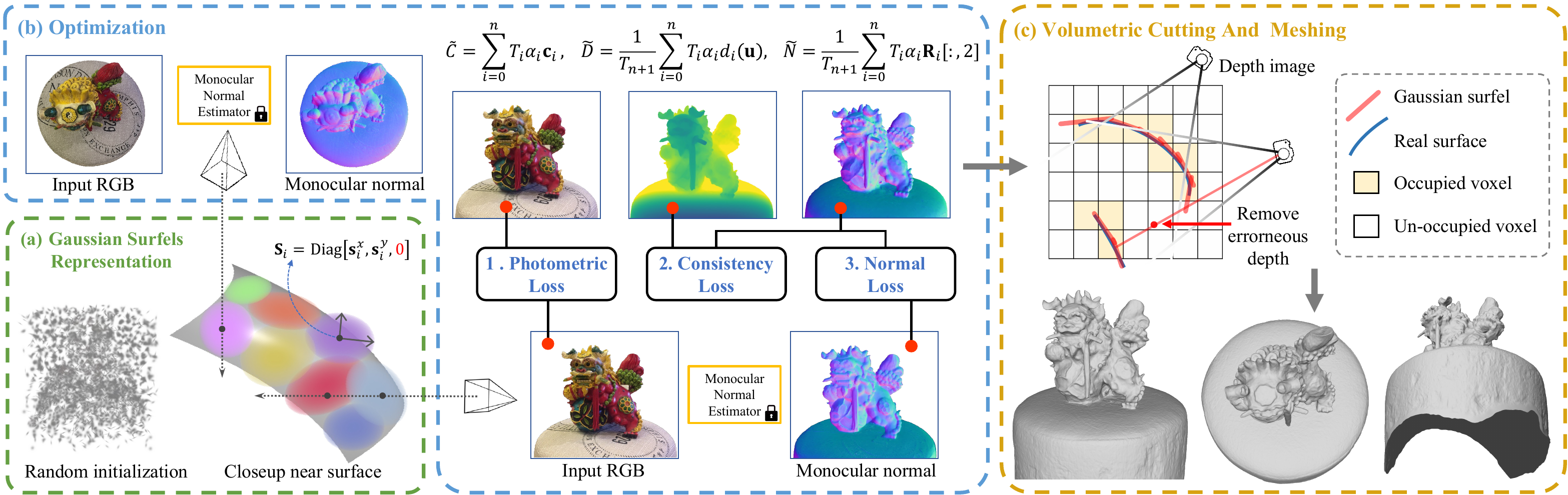}%
\caption{\textbf{The pipeline of our method.} \jiamin{Our method involves the following steps: \textcolor{green}{(a)} Starting with random initialization, our method represents the surface as a set of Gaussian surfels, each with learnable position, rotation, color, opacity, and covariance; \textcolor{blue}{(b)} Optimize the Gaussian surfels through multi-view photometric loss, depth-normal consistency loss, and normal prior loss; \textcolor{yellow}{(c)} Perform volumetric cutting on rendered depth maps, then apply Poisson meshing from rendered depth and normal to extract a high-quality mesh. Our method can automatically obtain an open surface reconstruction result.}
}
\label{fig:pipeline} 
\end{figure*}

However, NeRF and its variants extract isosurfaces based on heuristic thresholding of density values, which can introduce high-frequency noise into the reconstructed surfaces. To enhance the quality of reconstruction, several studies have suggested using occupancy grids \cite{oechsle2021unisurf} or signed distance functions (SDFs) \cite{yariv2020multiview,wang2021neus,yariv2021volume} represented by coordinate-based multi-layer perceptron (MLP) networks. These methods provide better-defined 3D surfaces compared to volume density fields. Follow-up works have further improved reconstruction by utilizing auxiliary information. For example, NeuralWarp \cite{darmon2022improving} employs patch warping with co-visibility information for surface optimization. Additionally, monocular depth and normals \cite{yu2022monosdf}, sparse point clouds \cite{fu2022geo, zhang2022critical}, semantic segmentation \cite{sun2022neural,guo2022neural}, and geometric priors \cite{long2022sparseneus} have been used to guide and regularize 3D reconstruction.

Alternatively, some methods incorporate explicit voxel grids \cite{wu2022voxurf} or multi-resolution hash encodings with a CUDA-based MLP implementation to expedite surface reconstruction \cite{zhao2022human,wang2023neus2,li2023neuralangelo}. For calculating the derivatives of Eikonal loss, Instant-NSR \cite{zhao2022human} approximates these derivatives using finite differences, while NeuS2 \cite{wang2023neus2} proposes a precise and efficient formulation of second-order derivatives tailored to MLPs. Neuralangelo \cite{li2023neuralangelo} also utilizes numerical gradients for computing higher-order derivatives as a smoothing operation. Additionally, it involves coarse-to-fine optimization on the hash grids, controlling different levels of detail.

\subsection{Neural point-based rendering.}

Point-based rendering has also been employed for neural rendering. Unlike volumetric representation, point-based methods represent geometry through unstructured samples in a topology-free manner~\cite{kobbelt2004survey}. As directly rendering point samples suffers from holes and discontinuities, surface splatting has been developed by splatting point primitives with an extent, such as ellipsoids or surfels with circular or elliptic discs~\cite{botsch2005high,pfister2000surfels,zwicker2002ewa,ren2002object,habbecke2007surface}, and the surfel representation has been applied to depth fusion in 3D reconstruction using depth cameras \cite{weise2009hand,xu2018online}. \jiamin{For point-based surface reconstruction, DSS~\cite{yifan2019differentiable} employs opaque circles to represent surfaces, optimizing them through differentiable surface splatting. This approach confines its gradients to few closest ellipses. Similarly, PBNR~\cite{kopanas2021point} uses an anisotropic 2D Gaussian associated with each pixel to capture local geometry. Due to the absence of adjustable opacity, both of these methods rely on a relatively precise geometric initialization.} Point-based rendering can also be combined with neural features and decoder networks to enhance rendering quality \cite{aliev2020neural,ruckert2022adop}.

Alternatively, recent methods have been developed to approximate volume rendering via conventional alpha-blending on sorted splats. \jiamin{Pulsar \cite{lassner2021pulsar} and Neural Catacaustics \cite{kopanas_neuralCatacaustics} represent surfaces as a set of isotropic 3D spheres or 2D Gaussian discs.} 3D Gaussian Splatting (3DGS) \cite{kerbl20233d} introduces a promising approach to modeling 3D scenes, achieving fast reconstruction and real-time rendering speeds, as well as enhanced quality. It represents complex scenes as a combination of numerous 3D Gaussians with position, opacity, anisotropic covariance, and spherical harmonic coefficients. SuGaR \cite{guedon2023sugar} further proposes a method allowing precise and fast mesh extraction from 3D Gaussian Splatting by adding a regularization term that encourages the Gaussians to align well with the surface. Similarly, NeuSG \cite{chen2023neusg} incorporates scale regularization to promote narrow 3D Gaussian ellipsoids. However, despite these efforts, achieving a close fit between anisotropic 3D Gaussians and surfaces remains a challenging task.


In contrast, our method takes a different approach by utilizing Gaussian surfels with opacity and anisotropic covariance to achieve a better alignment with the surface. Unlike 3DGS~\cite{kerbl20233d} and SuGaR~\cite{guedon2023sugar} which emphasize realistic rendering, our primary objective is to achieve super-fast surface reconstruction while preserving fine details.

\section{Method}
\subsection{Overview}

The goal of our method is to combine the flexible optimization and rendering scheme in 3DGS and the surface alignment property of surfels in a way such that surface meshes with fine details can be efficiently reconstructed. As outlined in Fig.~\ref{fig:pipeline}, the proposed method takes a set of posed RGB images as input, where $\mathbf{I}_k$ denotes the captured $k$-th image. As to the output, the method produces a set of Gaussian surfels, represented as ellipses  with anisotropic Gaussian kernels, opacities, and view-dependent colors represented using spherical harmonics (Sec. \ref{sec:rep}).

We optimize the Gaussian surfels to minimize the photometric difference between input RGB images and rendering results at captured views. To make the optimization controllable, we introduce a set of regularization losses to enforce surface smoothness and depth-normal consistency (Sec. \ref{sec:opt}). 
After completing the optimization, we render multi-view depth maps and normal maps and fuse them through screened Poison reconstruction \cite{kazhdan2013screened} to extract a high-quality global mesh. Before meshing, a volumetric cutting procedure is performed to reduce erroneous depth values caused by alpha blending for pixels at surface boundaries (see \ref{sec:meshing}).

\subsection{Gaussian Surfels}
\label{sec:rep}

In this section, we introduce the representation of our Gaussian surfels, which comprise a set of unstructured Gaussian kernels $\left\{ \mathbf{x}_i,\mathbf{r}_i,\mathbf{s}_i,o_i,\mathbf{\mathcal{C}}_i \right\}_{i\in \mathcal{P}}$, where $i$ is the index of each Gaussian kernel, $\mathbf{x}_i \in \mathbb{R}^3$ denotes the position of Gaussian kernel's center, $\mathbf{r}_i \in \mathbb{R}^4$ is its rotation (orientation) represented by a quaternion, $o_i \in \mathbb{R}$ is the opacity, and  $\mathbf{\mathcal{C}}_i \in \mathbb{R}^k$ is spherical harmonic coefficients of each Gaussian. The symbol $\mathbf{s}_i \in \mathbb{R}^3$ represents the scaling factors for each of the two local axes of a surfel. Thus, the 3D Gaussian distribution can be represented as:
\begin{align}
G\left( \mathbf{x};\mathbf{x}_i,\mathbf{\Sigma }_i \right) =\exp \left\{ -\small{0.5\left( \mathbf{x}-\mathbf{x}_i \right) ^{\top}}{\mathbf{\Sigma }_i}^{-1}\left( \mathbf{x}-\mathbf{x}_i \right) \right\},
\label{eq:gaussian_x}
\end{align}
where $\mathbf{\Sigma }_i$ is the covariance matrix expressed as the product of a scaling matrix $\mathbf{S}_i$ and a rotation matrix $\mathbf{R}(\mathbf{r}_i)$, which is a $3 \times 3$ rotation matrix represented by a quaternion $\mathbf{r}_i$.

We set $\mathbf{s}_i=[\mathbf{s}^x_i,\mathbf{s}^y_i,0]^\top$ to flatten the 3D Gaussians \cite{kerbl20233d}. Therefore, the corresponding $\mathbf{\Sigma }_i$ is changed to:
\begin{align}
\mathbf{\Sigma }_i=\mathbf{R}(\mathbf{r}_i)\mathbf{S}_i\mathbf{S}_{i}^{\top}\mathbf{R}(\mathbf{r}_i)^{\top}=\mathbf{R}(\mathbf{r}_i)\mathrm{Diag}\left[ \left( \mathbf{s}_{i}^{x} \right) ^2,\left( \mathbf{s}_{i}^{y} \right) ^2,0 \right] \mathbf{R}(\mathbf{r}_i)^{\top},
\label{equ:sigma}
\end{align}
where $\mathrm{Diag}[\cdot]$ indicates a diagonal matrix with diagonal entries in $[]$. In this Gaussian surfel representation, each Gaussian is truncated as a 2D ellipse and the normal for each Gaussian kernel can be directly computed as $\mathbf{n}_i = \mathbf{R}(\mathbf{r}_i)[:,2]$, where the $[\cdot]$ operator follows the slicing syntax of a multi-dimensional array in NumPy \cite{Kingma2015AdamAM}. Our method supports optimizing the normal direction of each ellipse to achieve the alignment of Gaussian surfels with the actual surface.

\paragraph{Differentiable Gaussian splatting}
For novel view rendering, the Gaussian splatting procedure remains consistent with that of 3DGS. Specifically, during rendering, the color of each pixel $\mathbf{u}$ is calculated via alpha-blending of all the nearby Gaussian kernels: 
\begin{align} \label{eq:render_rgb}
&\tilde{C}=\sum_{i=0}^n{T_i\alpha _i\mathbf{c}_i}, 
&T_i=\prod_{j=0}^{i-1}{(}1-\alpha _j),
&&\alpha_i=G'(\mathbf{u};\mathbf{u}_i,\mathbf{\Sigma }'_i)o_i,
\end{align}
where $\alpha_i$ represents alpha-blending weight, which is the product of opacity and the Gaussian weight specified in Eq.~\ref{eq:gaussian_x}. The view dependent color $\mathbf{c}_i$ is computed via spherical harmonics. To enhance the rendering efficiency, the 3D Gaussian in Eq.~\ref{equ:sigma} are reparameterized in 2D ray space \cite{zwicker2002ewa} as $G'$:
\begin{align}
G'\left( \mathbf{u};\mathbf{u}_i,\mathbf{\Sigma }'_i \right) &=\exp \left\{ -\small{0.5\left( \mathbf{u}-\mathbf{u}_i \right) ^{\top}}{\mathbf{\Sigma}'_i}^{-1}\left( \mathbf{u}-\mathbf{u}_i \right) \right\}, \\
\mathbf{\Sigma }'_i&=\left( \mathbf{J}_k\mathbf{W}_k\mathbf{\Sigma }_i\mathbf{W}_{k}^{\top}\mathbf{J}_{k}^{\top} \right) \left[:2,:2\right],
\end{align}
where $\mathbf{W}_k$ is a viewing transformation matrix for input image $k$, $\mathbf{J}_k$ is the affine approximation of the projective transformation. $\mathbf{\Sigma}'$ represents the transformed covariance matrix in image coordinates.

Similarly, the depth $\tilde{D}$ and normal $\tilde{N}$ for each pixel can also be calculated via Gaussian splatting and alpha-blending:
\begin{align} \label{eq:render_geom}
&\tilde{N}=\small{\frac{1}{1-T_{n+1}}}\sum_{i=0}^n{T_i\alpha _i\mathbf{R}_i\left[ :,2 \right]}, &\tilde{D}=\frac{1}{1-T_{n+1}}\sum_{i=0}^n{T_i\alpha _{i}d_i\left( \mathbf{u} \right)}.
\end{align}
We utilize ${1}/({1-T_{n+1}})$ to normalize the blending weight $T_i\alpha_i$. Unlike adding a background color during color rendering, we found this normalization approach to be more suitable for rendering depth and normal maps.

\begin{figure}[t]
  \centering
  \includegraphics[width=0.9\linewidth]{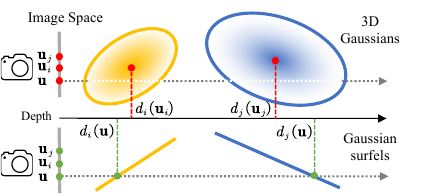}
  \caption{\emph{The first row}: the intersection of a ray with a 3D Gaussian point is challenging to calculate precisely. As a result, methods such as SuGaR~\cite{guedon2023sugar} approximate the depth of intersection with the depth of the Gaussian's center point, which can introduce errors. \emph{The second row}: the intersection of a ray with our Gaussian surfel can be calculated precisely, as well as its depth.}
  \label{fig: 3d_depth_ambig}
\end{figure}

For depth rendering, directly blending the depth of the center position $d_i(\mathbf{u}_i)$ of each Gaussian kernel is inaccurate because it neglects the slope of the 2D ellipse, as illustrated in Fig. \ref{fig: 3d_depth_ambig}. For instance, when dealing with Gaussian surfels aligned with a slanted plane, such depth rendering results will not be consistent with surface normals. To this end, in our implementation, the depth of pixel $\mathbf{u}$ for each Gaussian kernel $i$ is computed by calculating the intersection of the ray cast through pixel $\mathbf{u}$ with the Gaussian ellipse during splatting. It can be simplified via local Taylor expansion as:
\begin{align}
\label{eq:pix_depth}
d_i\left( \mathbf{u} \right) =d_i\left( \mathbf{u}_i \right) +\left( \mathbf{W}_k\mathbf{R}_i \right) \left[2,:\right]\mathbf{J}^{-1}_{pr}\left( \mathbf{u}-\mathbf{u}_i \right) ,
\end{align}
where $\mathbf{J}^{-1}_{pr}$ is the Jacobian of inverse mapping a pixel in the image space onto the tangent plane of Gaussain surfel as in~\cite{ZWicker2001}, $\left(\mathbf{W}_k\mathbf{R}_{i}\right)$ transforms the rotation matrix of a Gaussian surfel to the camera space. 

\subsection{Optimization}
\label{sec:opt}

To guide the optimization of explicit Gaussian surfels, our total loss $\mathcal{L}$ consists of five components: photometric loss $\mathcal{L}_{\mathrm{p}}$, normal-prior loss $\mathcal{L}_{\mathrm{n}}$, opacity loss $\mathcal{L}_{\mathrm{o}}$, depth-normal consistency loss $\mathcal{L}_{\mathrm{c}}$. Similar to~\cite{wang2021neus}, we also introduce a mask loss $\mathcal{L}_{\mathrm{m}}$, which computes binary cross-entropy between $\sum_{i=0}^n{T_i\alpha _i}$ and the binary segmentation mask. Thus we have:
\begin{align}
\mathcal{L}=
\mathcal{L}_{\mathrm{p}}+
\mathcal{L}_{\mathrm{n}}+
\lambda_{\mathrm{o}}\mathcal{L}_{\mathrm{o}}+
\lambda_{\mathrm{c}}\mathcal{L}_{\mathrm{c}}+
\lambda_{\mathrm{m}}\mathcal{L}_{\mathrm{m}},
\end{align}
where the trade-off weights $\lambda_{\mathrm{o}}$, $\lambda_{\mathrm{c}}$, and $\lambda_{\mathrm{m}}$ balance the loss terms. 

\paragraph{Photometric loss $\mathcal{L}_{\mathrm{p}}$} 

The photometric loss is same as in 3DGS. It combines an $L_1$ term and a D-SSIM term to minimize the difference between a rendered image $\tilde{\mathbf{I}}$ and its matched input image $\mathbf{I}$:
\begin{align} \label{eq:loss_photo}
\mathcal{L}_{\mathrm{p}}=0.8 \cdot {L}_1(\tilde{\mathbf{I}}, \mathbf{I})+0.2 \cdot {L}_{DSSIM}\cdot (\tilde{\mathbf{I}}, \mathbf{I}),
\end{align}

\paragraph{Depth-normal consistency loss $\mathcal{L}_{\mathrm{c}}$}

This term enforces consistency between the rendered depth $\tilde{\mathbf{D}}$ and the rendered normal $\tilde{\mathbf{N}}$:
\begin{align} \label{eq:loss_geom}
\mathcal{L}_{\mathrm{c}}=1-\tilde{\mathbf{N}}\cdot N(V(\tilde{\mathbf{D}})).
\end{align}
where $V(\cdot)$ transforms each pixel and its depth to a 3D point, and $N(\cdot)$ calculates the normal from neighboring points using the cross product. The depth-normal consistency loss plays a crucial role in our optimization process, particularly in resolving the gradient vanishing problem for each Gaussian surfel. 

\begin{figure}[t]
  \centering
  \includegraphics[width=1\linewidth]{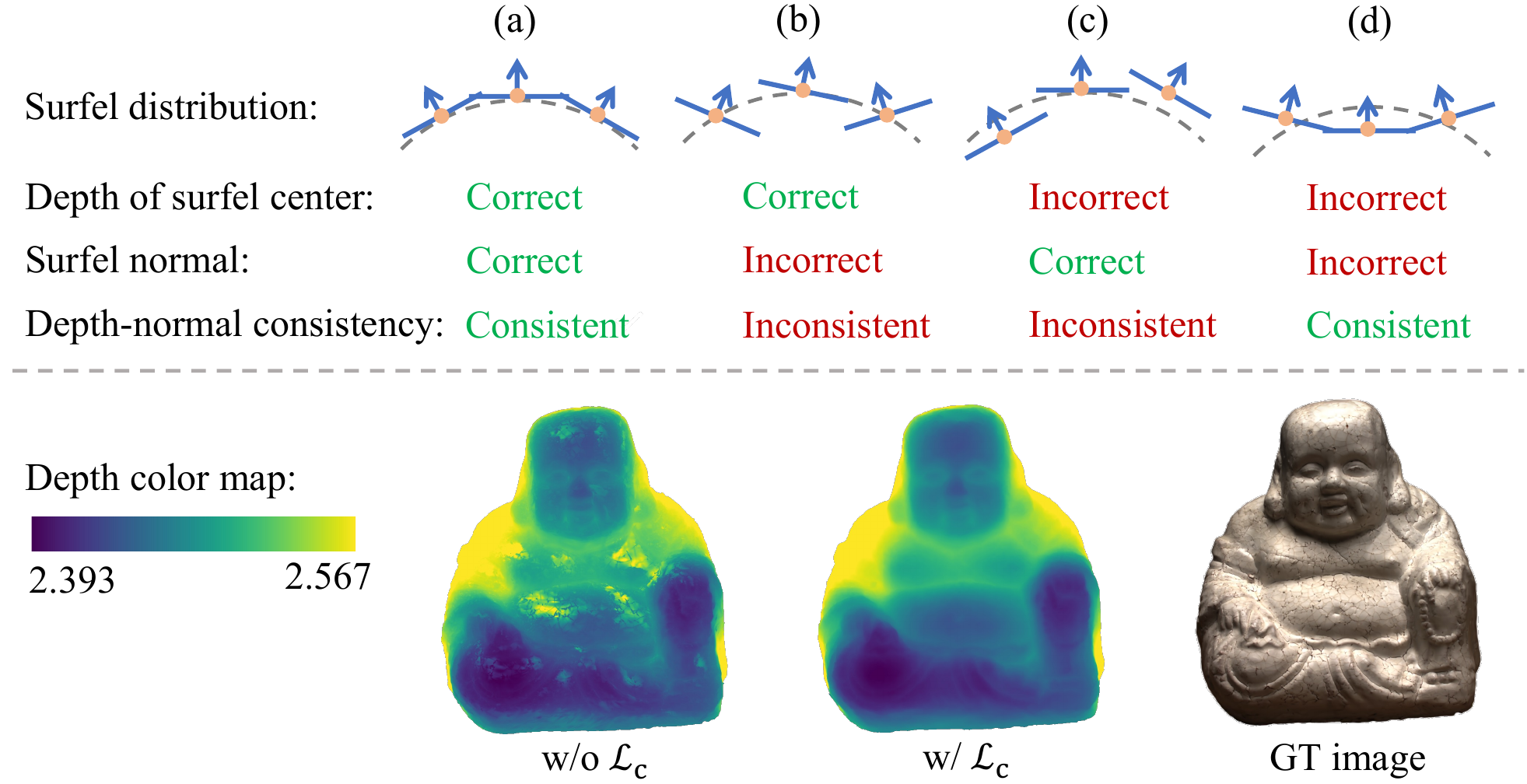}
  \caption {\jiamin{Depth-normal consistency. \emph{Top}: The consistency of surfel depth and normal during optimization. \emph{Bottom}: Rendered depth maps with and without consistency loss, alongside the reference image.}}
  \label{fig: depth_normal_cons}
\end{figure}

Moreover, we have found that it can also aid in resolving the ambiguity between rendered depth and normal, where Gaussian-splatted normals may exhibit accuracy while depths do not, and vice versa. 
As illustrated in Fig. \ref{fig: depth_normal_cons}: when the center depth is correct but the normal is not (top b), the rendered depth can help correct the direction of the Gaussian ellipse; when the normal is correct but the depth is not (top c), $\mathcal{L}_{\mathrm{c}}$ can be interpreted as normal-aware depth smoothing. Enforcing bidirectional consistency between rendered depth and normal results in an improved quality of reconstruction in both cases. 
\jiamin{The self-supervised consistency loss alone does not guarantee correct surface (top d), while combining with other loss terms in this section achieves consistent and correct results (top a).}

\paragraph{Normal-prior loss $\mathcal{L}_{\mathrm{n}}$} 
This term acts as a prior-based regularizer that improve optimization stability, especially in areas with highlights, where photometric loss may result in wrong surface points. We use normal maps $\hat{N}$ from a pretrained monocular deep neural network from Omnidata~\cite{Ainaz2021Omnidata}. Additionally, we introduce an $L_1$ loss to minimize the gradient of the rendered normal, denoted as $\nabla{\tilde{N}}$, thereby regularizing the curvature of the surface:
\begin{align} \label{eq:loss_normal}
\mathcal{L}_{\mathrm{n}}=
0.04 \cdot (1-\tilde{\mathbf{N}}\cdot \hat{\mathbf{N}})+
0.005 \cdot L_1(\nabla{\tilde{\mathbf{N}}}, \mathbf{0}),
\end{align}


\paragraph{Opacity loss $\mathcal{L}_{\mathrm{o}}$}

This opacity loss promotes non-transparent surfaces by encouraging each Gaussian's opacity to be either near zero or near one, where $o_i$ is parameterized with a sigmoid function. This contributes to the overall quality of the reconstruction:
\begin{align} \label{eq:loss_solid}
\mathcal{L} _{\mathrm{o}}=\exp \left( {-(o_i-0.5)^2}/{0.05} \right) .
\end{align}



According to Equation \ref{equ:sigma}, the gradient of the covariance matrix with respect to the 3rd column of $\mathbf{R}_i$ (normal), is equal to zero. Consequently, following the chain rule, the photometric loss $\mathcal{L}_\mathrm{p}$ has no gradient with respect to the normal of each Gaussian surfel. As $\mathbf{R}_i$ is a rotation matrix in $\mathbb{SO}(3)$, the normal will still be modified according to the first two axes of $\mathbf{R}_i$. However, this kind of indirect modification can lead to errors during optimization. This observation inspired us to incorporate the depth-normal consistency loss $\mathcal{L}_\mathrm{c}$, which rectifies the normal of each Gaussian surfel using the gradient obtained from the rendered depth.



\subsection{Gaussian Point Cutting and Meshing}
\label{sec:meshing}
We fuse the rendered depth and normal maps at captured views and then apply screened Poisson reconstruction~\cite{kazhdan2013screened} with a tree depth of 10 to obtain the final surface mesh. Compared with directly applying Poisson reconstruction to Gaussian centers, this greatly increases the point density and improves the quality of surface details. 

\begin{figure}[t]
  \centering
  \includegraphics[width=1\linewidth]{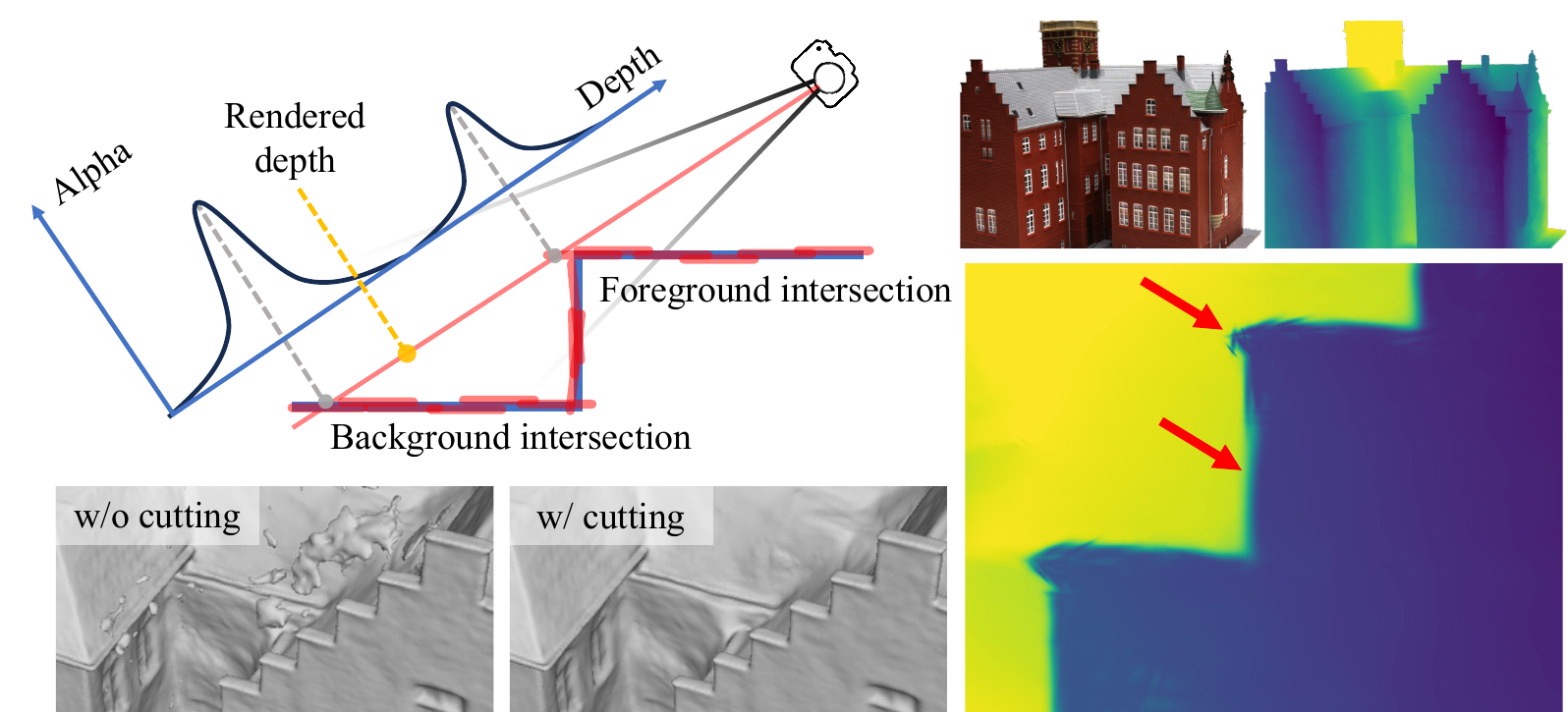}
  \caption{An example of error in the rendered depth. The complex distribution of Gaussian surfels after optimization makes it difficult to remove outlier Gaussian points along each ray by discarding points far from median or alpha-weighted mean.}
  \vspace{-4mm}
  \label{fig:depth_blending}
\end{figure}

\begin{figure}[t]
  \centering
  \includegraphics[width=0.99\linewidth]{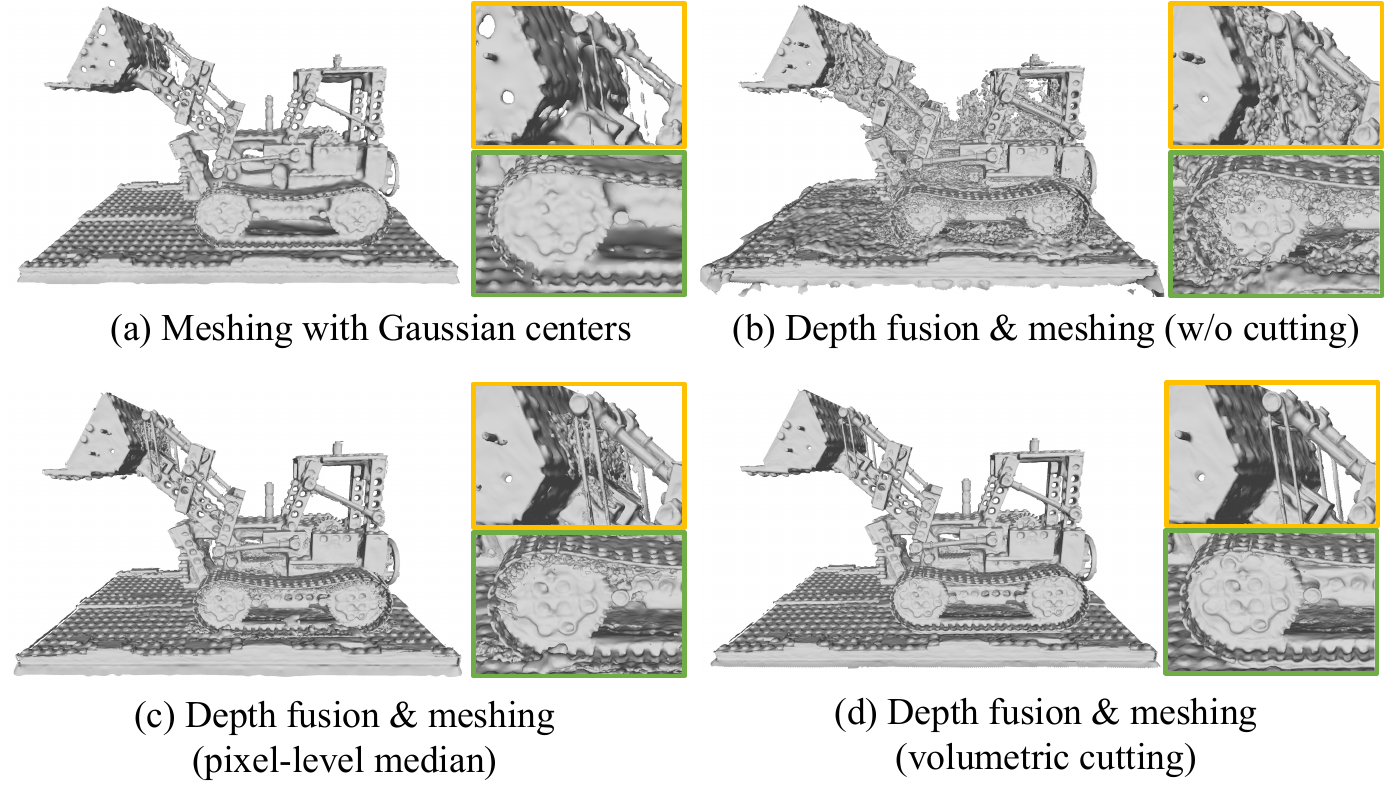}
  \caption {Comparisons on cutting strategies. Meshing is achieved via screened Poisson reconstruction \cite{kazhdan2013screened}.}
  \label{fig:volumetric_cutting}
\end{figure}

However, the rendered depth map still contains errors, especially near depth discontinuities. As depicted in Fig. \ref{fig:depth_blending}, if ellipses extend beyond the true surface boundary and the associated weights cannot rapidly decay to zero, the rendered depth of the background surfaces will be influenced by the alpha of the foreground Gaussian points, resulting in a depth that is in front of the true background surface. Due to the complex distribution of Gaussian surfels, we found it was difficult to remove outliers along each ray by discarding surfels far from median or alpha-weighted mean (Fig.~\ref{fig:volumetric_cutting} (c)). Instead, we implement volumetric cutting according to the aggregated alpha values from each Gaussian surfel.

\paragraph{Volumetric cutting}  The idea is to mark those voxels far from Gaussian surfels as un-occupied, i.e., cutting them from the grid. Thus, the wrong 3D point in red illustrated in Fig.~\ref{fig:depth_blending} can be removed because it is inside a un-occupied voxel. More precisely, we first construct a $512^3$ voxel grid within the bounding box. Next, we traverse all the Gaussian ellipses, calculate their intersection with the surrounding voxels, and accumulate the weighted opacity $G( \mathbf{x};\mathbf{x}_i,\mathbf{\Sigma }_i) \cdot o_i$ to the corresponding voxels. To reduce computational cost, we approximate the integration of Gaussian weights and opacities within the intersection area using the weighted opacity of the voxel center. If a voxel has a low accumulated weighted opacity, lower than $\lambda=1$ in our experiments, indicating a large distance from the foreground or background surfaces, we prune these voxels as well as the 3D points in them computed from the depth. As shown in Fig. \ref{fig:volumetric_cutting}, compared with outlier removal using median depth at each pixel, our volumetric cutting exhibits better quality and computational efficiency.



\subsection{Implementation details}

\begin{figure}[t]
  \centering
  \includegraphics[width=1.0\linewidth]{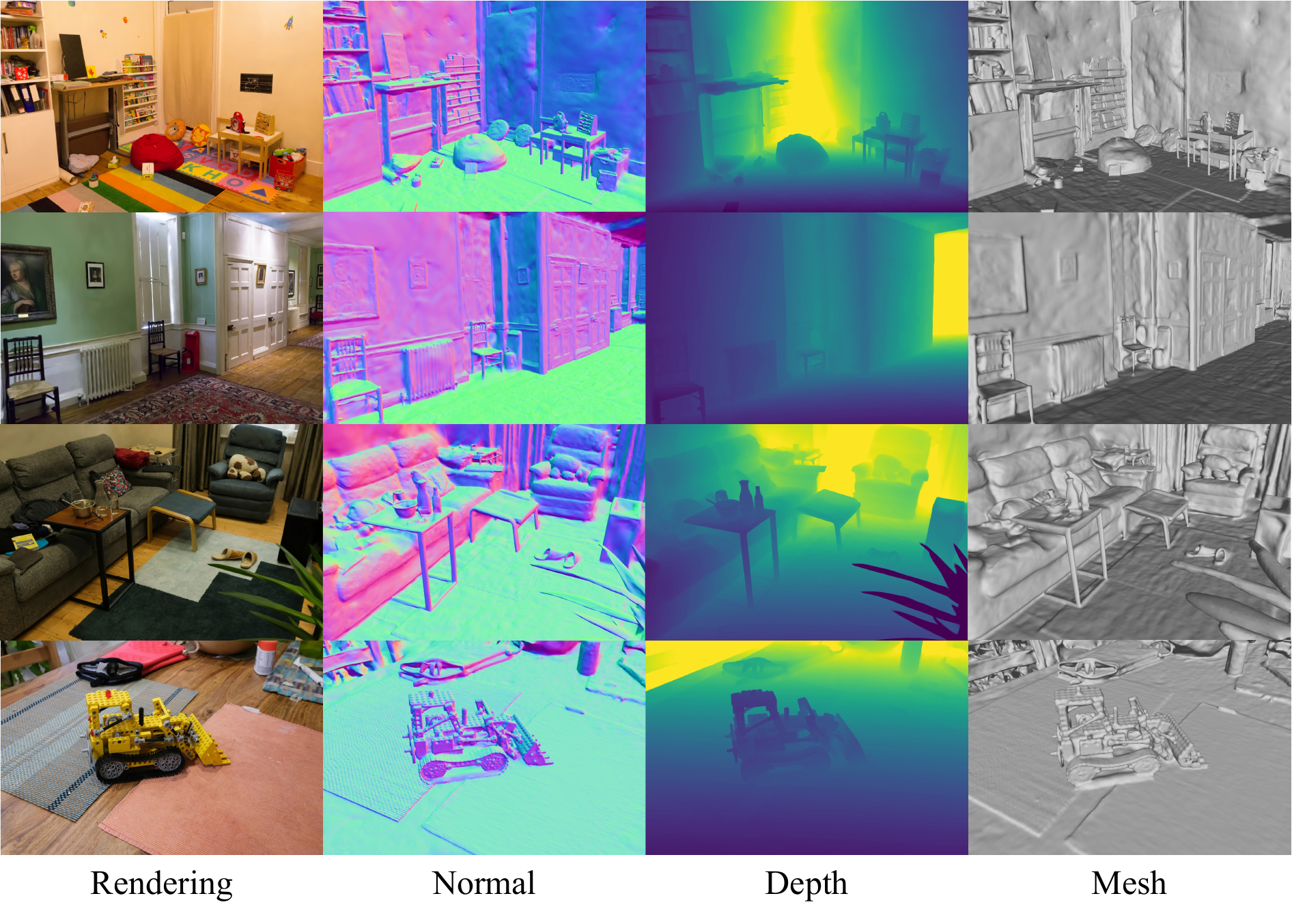}
  \caption{Reconstruction results of our method on indoor scenes from DeepBlending \cite{DeepBlending2018} and MipNeRF360 \cite{barron2022mipnerf360} datasets when initialized with sparse SfM points.}
  \label{fig:big_scene}
\end{figure}

\paragraph{Initialization} We support the use of sparse points computed via the Structure from Motion (SfM) as the initialization (Fig. ~\ref{fig:big_scene}). We found this could accelerate the decrease of loss function values in the first few steps but overall did not significantly improve the convergence rate. Thus, we initialize our Gaussian surfels with random positions and rotations inside the approximated bounding box of the target object in all our experiments.

\paragraph{Optimization}

Our model is trained on a GPU server with an i9-14900K CPU and a RTX 4090 GPU, using the Adam optimizer~\cite{Kingma2015AdamAM}, PyTorch 1.12.0~\cite{pytorch}, and CUDA 11.8. Each data batch contains all pixels of an image at every iteration. We do the ADMM training procedure for 15k iterations starting at lower resolutions for a warm-up. For the training loss, we assign $\lambda_\mathrm{o}=0.01$ and $\lambda_\mathrm{m}=1$. $\lambda_\mathrm{c}$ is linearly increased from $0$ to $0.1$. The learning rate associated with the surfels position $\mathbf{x}_i$ is set to 1.6e-4, and decays exponentially to 1.6e-6. The learning rates for $\mathbf{r}_i, \mathbf{s}_i, o_i, \mathcal{C}_i$ are set to 1.0e-3, 5.0e-3, 5.0e-2, 2.5e-3, respectively. As the photometric loss does not have a gradient with respect to $\mathbf{R}(\mathbf{r}_i)$[:,2], we scale the gradient $\partial \tilde{\mathbf{N}}/\partial \mathbf{R}[:,2]$ by a factor of 10 in the derivative chain to balance the gradients along each axis in $\mathbf{R}(\mathbf{r}_i)$, and guide the Gaussian ellipse to rotate to the correct direction. 

We utilize adaptive point splitting, cloning, and pruning techniques akin to 3DGS. A novel aspect of our implementation involves pruning points that do not receive gradients every N iterations, where N is the number of images. This pruning is motivated by the fact that points lacking gradients correspond to those that are invisible from all views, such as noisy points within the object.

\section{Experiments}

\paragraph{Baseline} 
We compare our method with 1) NeuS \cite{wang2021neus}, INSR \cite{zhao2022human} and NeuS2 \cite{wang2023neus2}, which are implicit surface reconstructions by neural volume rendering, 2) 3DGS \cite{kerbl20233d} and SuGaR \cite{guedon2023sugar} which are 3D Gaussian points-based surface reconstructions. All comparisons use original author's implementations and hyperparameters.

\paragraph{Evaluation metrics and datasets}
To evaluate our method, we report both surface accuracy as Chamfer distance and render fidelity as Peak Signal-to-Noise Ratio (PSNR) on the DTU \cite{jensen2014large} and BlendedMVS \cite{yao2020blendedmvs} datasets. \jiamin{We follow NeuS2 to leave $7\sim 8$ images for testing on DTU, and use all images for training on BlendedMVS.} The Chamfer distance measures the average of accuracy and completeness, i.e. the Chamfer distance of prediction to reference point cloud and vise versa.

\begin{figure*}[]
\centering
\includegraphics[width=\linewidth]{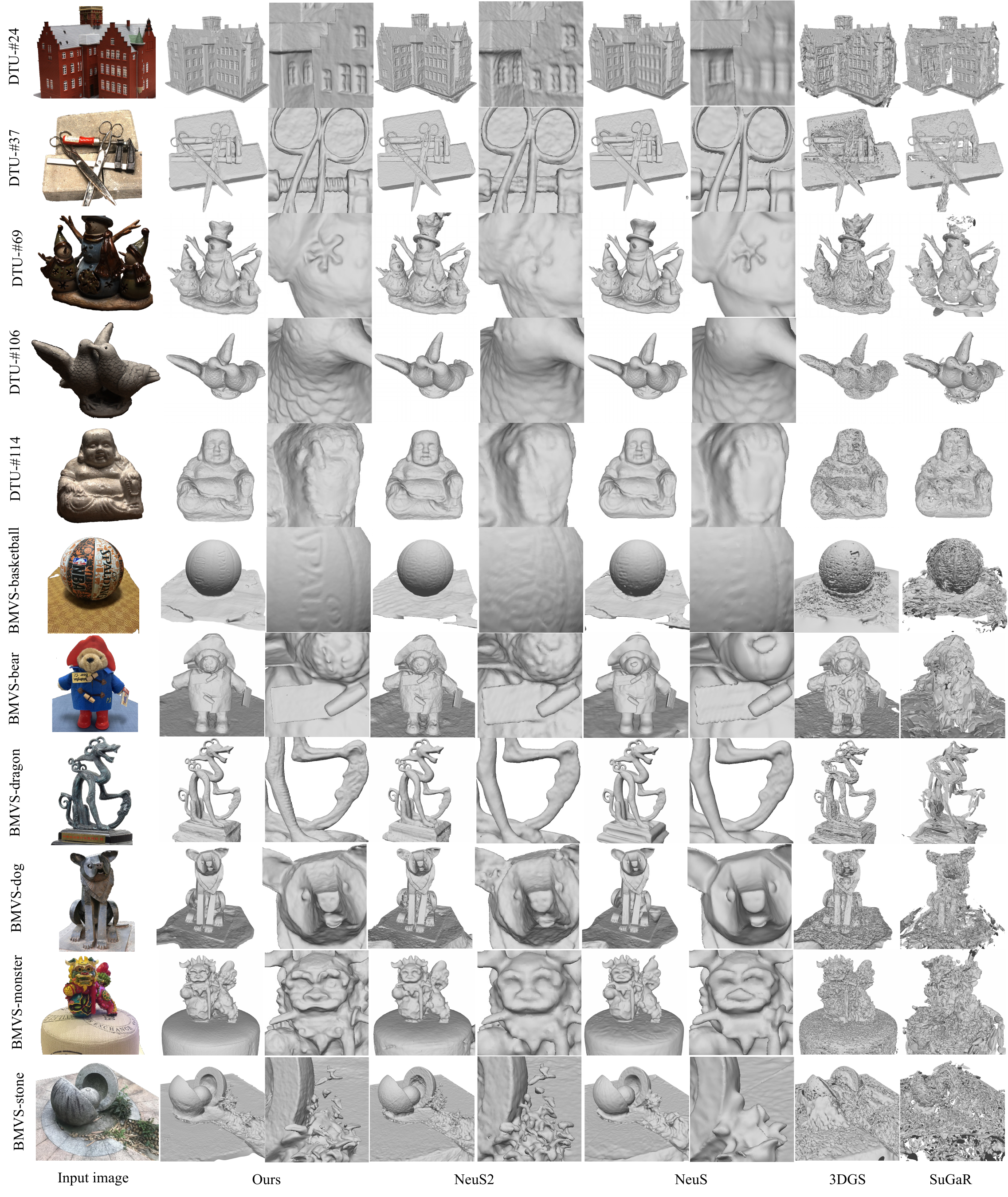}%
\caption{\textbf{Qualitative comparisons on DTU and BlendedMVS Datasets.} 
}
\label{fig:comparison} 
\end{figure*}

\begin{table}[t!]
\centering
\caption{Geometry quality comparison on the DTU dataset. We report the Chamfer distance compared with baselines. Mean scores of Chamfer distance (mm) and training times (minute) are included on the bottom. Best results are highlighted as \colorbox{colorFst}{1st}, \colorbox{colorSnd}{2nd}, and \colorbox{colorTrd}{3rd}. }\label{tab:cmp_dtu}
\begin{tabular}{c|cccccc}
\toprule
Methods & Ours & 3DGS & SuGaR & NeuS & NeuS2 & INSR \\ \hline
24 & \colorbox{colorSnd}{0.66} & 2.89 & 1.85 &  \colorbox{colorTrd}{0.83} & \colorbox{colorFst}{0.56} & 2.86\\
37 & \colorbox{colorSnd}{0.93} & 2.65 & {1.19} & \colorbox{colorTrd}{0.98} & \colorbox{colorFst}{0.76} & 2.81 \\
40 & \colorbox{colorSnd}{0.54} & 2.66 & 1.91 & \colorbox{colorTrd}{0.56} & \colorbox{colorFst}{0.49} & 2.09\\
55 & \colorbox{colorSnd}{0.41} & 2.31 & 1.64 & \colorbox{colorFst}{0.37} & \colorbox{colorFst}{0.37} & 0.81\\
63 & \colorbox{colorSnd}{1.06} & 3.55 & 2.76 & \colorbox{colorTrd}{1.13} & \colorbox{colorFst}{0.92} & 1.65\\
65 & \colorbox{colorTrd}{1.14} & 2.94 & 1.94 & \colorbox{colorFst}{0.59} & \colorbox{colorSnd}{0.71} & 1.39\\
69 & \colorbox{colorTrd}{0.85} & 2.19 & 1.97 & \colorbox{colorFst}{0.60} & \colorbox{colorSnd}{0.76} & 1.47\\
83 & \colorbox{colorSnd}{1.29} & 2.66 & {3.07} & \colorbox{colorTrd}{1.45} & \colorbox{colorFst}{1.22} & 1.67\\
97 & \colorbox{colorTrd}{1.53} & 2.99 & 2.38 & \colorbox{colorFst}{0.95} & \colorbox{colorSnd}{1.08} & 2.47\\
105 & \colorbox{colorTrd}{0.79} & 1.83 & 1.33 & \colorbox{colorSnd}{0.78} & \colorbox{colorFst}{0.63} & 1.12\\
106 & \colorbox{colorTrd}{0.82} & 2.54 & 1.69 & \colorbox{colorFst}{0.52} & \colorbox{colorSnd}{0.59} & 1.22\\
110 & \colorbox{colorTrd}{1.58} & 3.38 & 3.61 & \colorbox{colorSnd}{1.43} & \colorbox{colorFst}{0.89} & 2.30\\
114 & \colorbox{colorTrd}{0.45} & 2.05 & 1.53 & \colorbox{colorFst}{0.36} & \colorbox{colorSnd}{0.40} & 0.98\\
118 & \colorbox{colorTrd}{0.66} & 1.99 & 1.83 & \colorbox{colorFst}{0.45} & \colorbox{colorSnd}{0.48} & 1.41\\
122 & \colorbox{colorSnd}{0.53} & 2.02 & 1.97 & \colorbox{colorFst}{0.45} & \colorbox{colorTrd}{0.55} & 0.95\\ \hline
Mean & \colorbox{colorTrd}{0.88} & 2.58 & 2.05 & \colorbox{colorSnd}{0.77} & \colorbox{colorFst}{0.70} & 1.68\\
Time & \colorbox{colorTrd}{6.67} & \colorbox{colorSnd}{5.19} & 30.9 & 408 & \colorbox{colorFst}{3.27} & 8.48\\
\bottomrule
\end{tabular}
\end{table}
\begin{table}[t]
\centering
\caption{Geometry quality comparison on the BlendedMVS dataset. Best results are highlighted as \colorbox{colorFst}{1st}, \colorbox{colorSnd}{2nd}, and \colorbox{colorTrd}{3rd}.}\label{tab:cmp_bmvs}
\begin{tabular}{c|cccccc}
\toprule
Methods & Ours & 3DGS & SuGaR & NeuS & NeuS2 & INSR \\ \hline
Basketball  & \colorbox{colorFst}{1.55} & {5.59} & {8.00} & {2.96} & \colorbox{colorSnd}{2.48} & \colorbox{colorTrd}{2.67} \\
Bear        & \colorbox{colorFst}{1.96} & {5.08} & {9.73} & \colorbox{colorSnd}{3.00} & \colorbox{colorTrd}{3.20} & {3.72} \\
Bread       & \colorbox{colorFst}{1.17} & {3.98} & {7.65} & {2.85} & \colorbox{colorSnd}{2.47} & \colorbox{colorTrd}{2.76} \\
Camera      & \colorbox{colorFst}{2.34} & {5.85} & {7.77} & \colorbox{colorSnd}{2.61} & \colorbox{colorTrd}{2.89} & {3.25} \\
Clock       & \colorbox{colorTrd}{3.41} & {7.34} & {9.21} & \colorbox{colorFst}{2.75} & \colorbox{colorSnd}{2.87} & {3.45} \\
Cow         & \colorbox{colorTrd}{2.17} & {5.38} & {8.69} & \colorbox{colorFst}{2.04} & {2.29} & \colorbox{colorSnd}{2.11} \\
Dog         & \colorbox{colorTrd}{2.89} & {7.16} & {9.27} & \colorbox{colorSnd}{2.75} & {2.97} & \colorbox{colorFst}{2.54} \\
Doll        & \colorbox{colorFst}{2.13} & {6.25} & {8.75} & \colorbox{colorSnd}{2.17} & \colorbox{colorTrd}{2.23} & {2.44} \\
Dragon      & \colorbox{colorTrd}{2.75} & {4.90} & {9.76} & {2.95} & \colorbox{colorSnd}{2.68} & \colorbox{colorFst}{2.20} \\
Durian      & \colorbox{colorFst}{2.35} & {6.66} & {8.04} & \colorbox{colorSnd}{3.14} & \colorbox{colorTrd}{3.39} & {5.78} \\
Fountain    & \colorbox{colorSnd}{3.01} & {5.03} & {9.05} & \colorbox{colorTrd}{3.03} & \colorbox{colorFst}{2.68} & {3.79} \\
Gundam      & \colorbox{colorFst}{0.85} & {4.46} & {7.32} & \colorbox{colorTrd}{1.62} & {1.87} & \colorbox{colorSnd}{1.36} \\
House       & \colorbox{colorFst}{2.12} & {4.64} & {7.28} & {3.23} & \colorbox{colorTrd}{3.02} & \colorbox{colorSnd}{2.93} \\
Jade        & \colorbox{colorFst}{3.50} & {6.96} & {10.7} & {4.25} & \colorbox{colorTrd}{4.03} & \colorbox{colorSnd}{3.85} \\ 
Man         & {2.47} & {7.13} & {9.29} & \colorbox{colorTrd}{2.29} & \colorbox{colorFst}{2.20} & \colorbox{colorSnd}{2.22} \\
Monster     & \colorbox{colorFst}{1.38} & {5.84} & {8.20} & \colorbox{colorTrd}{1.92} & {1.96} & \colorbox{colorSnd}{1.80} \\
Sculpture   & {2.90} & {6.68} & {8.98} & \colorbox{colorTrd}{2.10} & \colorbox{colorSnd}{2.06} & \colorbox{colorFst}{1.70} \\
Stone       & \colorbox{colorFst}{1.84} & {6.01} & {9.19} & {2.51} & \colorbox{colorSnd}{2.01} & \colorbox{colorTrd}{2.12} \\
\hline
Mean        & \colorbox{colorFst}{2.27} & 5.83 & 8.71 & \colorbox{colorTrd}{2.67} & \colorbox{colorSnd}{2.63} & 2.82\\
Time & {3.82} & \colorbox{colorSnd}{3.47} & 18.3 & {377} & \colorbox{colorTrd}{3.61} & \colorbox{colorFst}{2.78}\\
\bottomrule
\end{tabular}
\end{table}
\paragraph{Comparisons} We first benchmark our method on the DTU dataset \cite{jensen2014large} both quantitatively (Table \ref{tab:cmp_dtu}) and qualitatively (Fig. \ref{fig:comparison}). The dataset encompasses laboratory-captured scenes, each encompassing 49 or 64 images with resolution $1600 \times 1200$, and we choose the same set of scenes selected by IDR \cite{yariv2020multiview} that contain 15 objects with manually annotated object masks. We further tested on 18 challenging scenes from the low-res set of the BlendedMVS dataset~\cite{yao2020blendedmvs}, where each scene comprises 24 to 143 images at $768 \times 576$ pixels with masks provided. The evaluation result is obtained using the DTU evaluation code.

The statistics of Chamfer distances in these two tests are shown in Tables \ref{tab:cmp_dtu} and \ref{tab:cmp_bmvs}. Our method significantly outperforms 3DGS and SuGaR on the DTU and the BlendedMVS datasets. \jiamin{Compared with NeuS2, our results show a larger Chamfer distance on DTU. We hypothesize that it is due to the bias inherent in per-view depth calculation using alpha blending. Despite the higher Chamfer distance, our results often have less noise than NeuS2 and more details than NeuS, as illustrated in Fig.~\ref{fig:comparison}.} In addition, our method can be applied to the reconstruction of open surfaces (Fig.~\ref{fig:pipeline}), since Gaussian surfels do not assume closed surfaces as in the representation of signed distance functions. 
Overall, our method achieves a good balance between reconstruction speed and quality. In contrast to NeuS, our approach exhibits rapid convergence to high-quality reconstructions, as detailed in Table \ref{tab:cmp_dtu} and \ref{tab:cmp_bmvs}. Furthermore, while NeuS \cite{wang2021neus} employs a large MLP to model surfaces, offering robustness to outliers, it may result in over-smoothing of the reconstructed surfaces
as depicted in Fig.~\ref{fig:comparison}. On the other hand, our reconstruction speed is comparable to that of INSR \cite{zhao2022human} and NeuS2 \cite{wang2023neus2}, both of which utilize hash feature maps and tiny MLPs to expedite implicit surface optimization.

Compared to point-based reconstruction methods such as 3DGS \cite{kerbl20233d} and SuGaR~\cite{guedon2023sugar}, both of which utilize Gaussian points to represent radiance fields and surfaces, our method excels in reconstructing noise-free surfaces and capturing intricate details. For 3DGS, we employ a mesh extraction method similar to ours but calculate normals directly based on the rendered depth map. For SuGaR, we utilize the proposed $\lambda-$level set of the density function with $\lambda = 0.3$ for point extraction and Poisson reconstruction with a depth of 10. As illustrated in Fig. \ref{fig:comparison}, while superior to vanilla 3DGS on DTU, this approach still exhibits ellipsoid-like artifacts and holes on the surface. We believe this discrepancy arises because the regularization term that encourages the 3D Gaussians to be flat does not align well enough with the extracted $\lambda-$level set. 
\begin{table}[t]
\centering
\small
\caption{Rendering quality comparison on DTU in common (12.5\% of images for testing) and sparse (50\% for testing) settings.}
\label{tab:cmp_sparse}
\begin{tabular}{c|ccc|ccc}    
\toprule
    Test ratio & \multicolumn{3}{c|}{12.5\%} & \multicolumn{3}{c}{50\%} \\ 
    Methods & Ours & 3DGS & NeuS2 & Ours & 3DGS & NeuS2 \\ \hline
    PSNR $\uparrow$ & \colorbox{colorSnd}{32.51} & \colorbox{colorFst}{32.78} & 31.41 & \colorbox{colorFst}{31.70} & 30.08 & \colorbox{colorSnd}{30.66} \\
    SSIM $\uparrow$ & \colorbox{colorSnd}{0.942} & \colorbox{colorFst}{0.943} & 0.916 & \colorbox{colorFst}{0.936} & 0.882 & \colorbox{colorSnd}{0.910}\\
\bottomrule
\end{tabular}
\end{table}
\begin{figure}[t]
  \centering
  \includegraphics[width=1.0\linewidth]{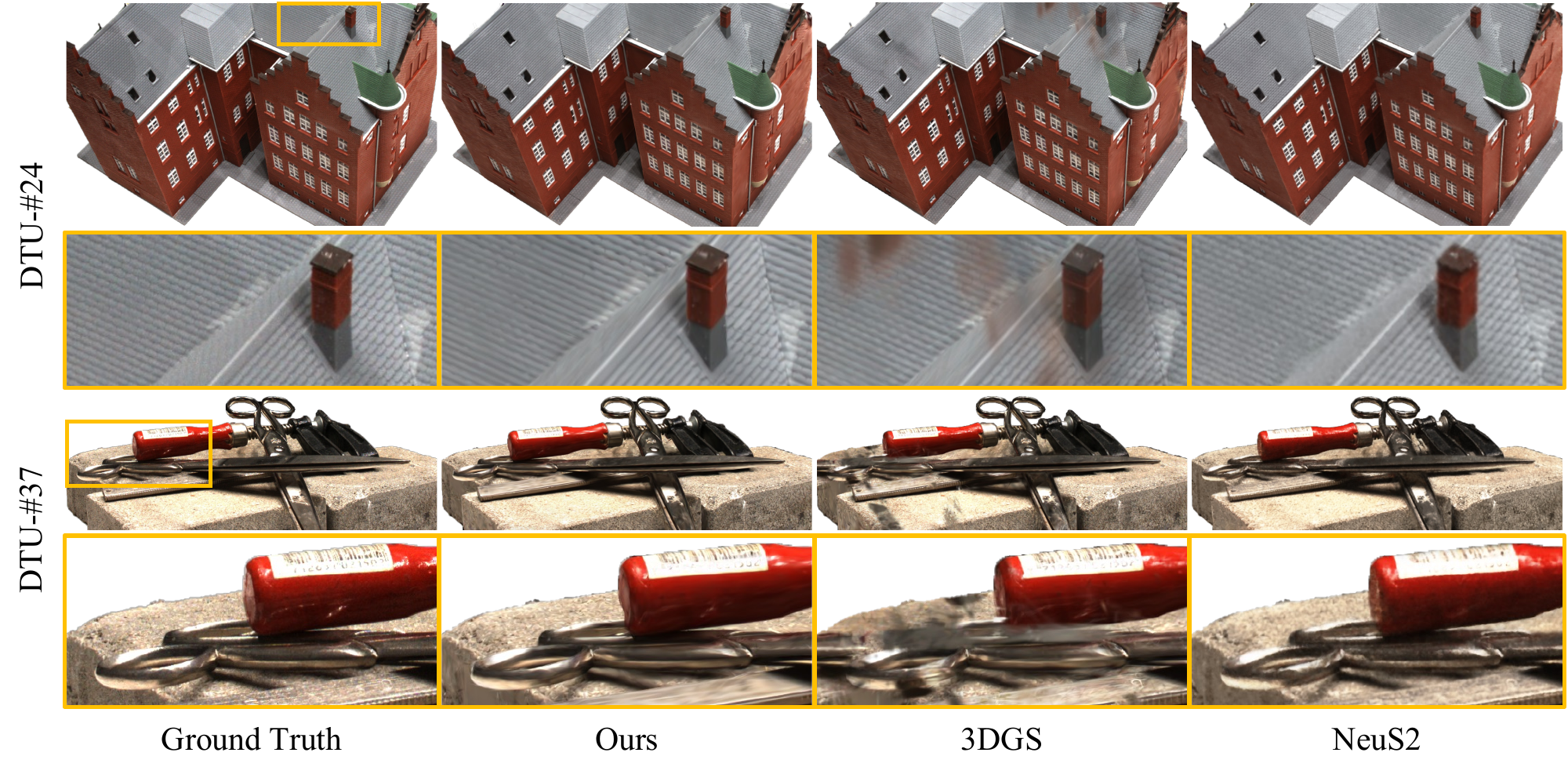}
  \caption{Rendering quality comparison under sparse inputs on DTU. With geometry constraints, our method and NeuS2 render floater-free images compared to 3DGS. Meanwhile our renderings recover better visual details than NeuS2's.}
  \label{fig:sparse}
\end{figure}

\jiamin{We also assess the rendering quality of our method by comparing it with 3DGS and NeuS2 on DTU under two settings: a common setting where 12.5\% of images are reserved for testing, and a sparse setting where half of the images are reserved for testing. As shown in Table~\ref{tab:cmp_sparse}, our method exceeds NeuS2 in both settings, only performs slightly worse than the vanilla 3DGS in common setting. However, due to its precise underlying geometry, our method demonstrates superior generality compared to 3DGS, resulting in a significant improvement in rendering quality in sparse settings (Fig.~\ref{fig:sparse}).}

\begin{table}[t]
\setlength{\tabcolsep}{4.5pt}
\centering
\small
\caption{Loss term ablation studies on DTU using average Chamfer distance (CD) and PSNR score.}\label{tab:abl_dtu}
\begin{tabular}{c|cccccc}
\toprule
Metrics & Full & w/o $\mathcal{L}_\mathrm{c}$ & w/o $\mathcal{L}_\mathrm{n}$ & w/o $\mathcal{L}_\mathrm{o}$ & w/o $\mathcal{L}_\mathrm{m}$ & w/o cut\\
\hline
CD $\downarrow$ & \colorbox{colorFst}{0.882} & 1.243 & \colorbox{colorTrd}{1.070} & 1.085 & \colorbox{colorSnd}{1.015} & 1.189 \\
PSNR $\uparrow$ & \colorbox{colorSnd}{32.51} & 31.56 & \colorbox{colorFst}{32.63} & \colorbox{colorTrd}{32.08} & 29.10 & /\\
\bottomrule
\end{tabular}
\end{table}
\begin{figure}[]
  \centering
  \includegraphics[width=1.0\linewidth]{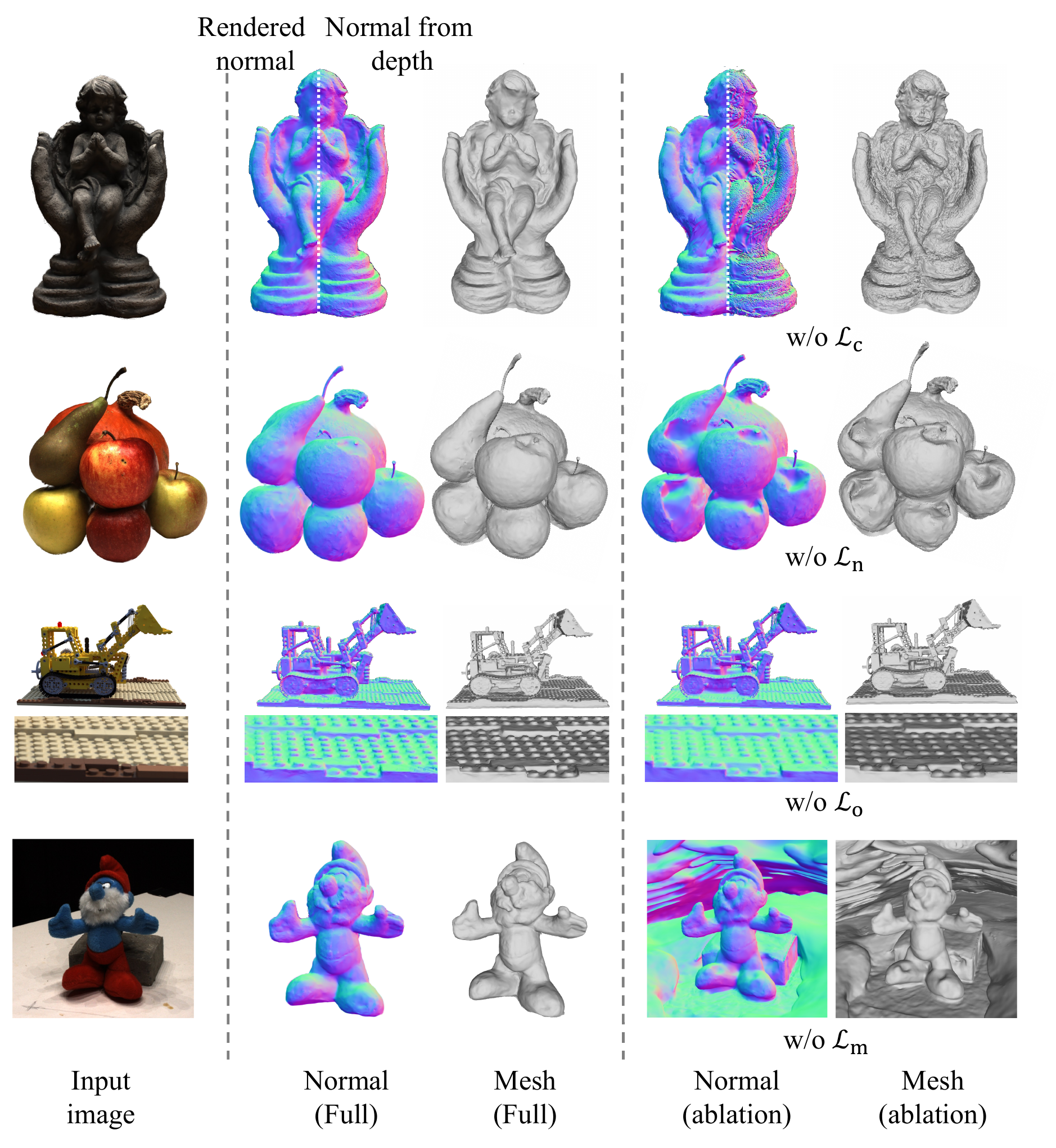}
  \caption{\emph{Top}: Ablation on depth-normal consistency loss. For the normal column, the part left to the central dashed line on the object represents the rendered normals, while the right part represents the normals calculated from the rendered depth. We can observe better consistency with the depth-normal consistency loss, resulting in improved reconstruction. \emph{Second row}: Ablation on normal-prior loss; the concave parts on the surface caused by highlights are corrected. \emph{Third row}: Ablation on opacity loss. The LEGO studs are better reconstructed, with sharper edges. \emph{Bottom} Ablation without masks, where the absence of masks may lead to noises in texture-less background regions.}
  \label{fig:abl}
\end{figure}
\paragraph{Ablations}
We show quantitative analyses of our loss terms by excluding each individually from the optimization (Table \ref{tab:abl_dtu}). The photometric loss $\mathcal{L}_\mathrm{p}$ is always included. In addition, we evaluate the effect of volumetric cutting by excluding volumetric cutting (w/o cut). Removing depth-normal consistency loss $\mathcal{L}_\mathrm{c}$ significantly reduces reconstruction quality. The qualitative results of the ablation are shown in Fig.~\ref{fig:abl}. The normal-prior loss helps reduce fluctuations in the geometric reconstruction in highlighted areas (the second row of Fig.~\ref{fig:abl}), contributing to a lower Chamfer distance. However, as indicated in the third colomn of Table~\ref{tab:abl_dtu}, removing the normal-prior loss can slightly improve the rendering quality. This is because this loss might prevent the optimizer from approximating the highlights as a virtual light sources behind the actual surface, resulting in lower rendering quality. The opacity loss aids in representing details (the fourth column in~\ref{tab:abl_dtu}). As depicted in the third rows of Fig.~\ref{fig:abl}, with this loss, the details on the ``LEGO studs'' are much more clearly reconstructed and rendered. 
The last two columns of Table~\ref{tab:abl_dtu} demonstrate the efficacy of mask loss $\mathcal{L}_\mathrm{m}$ and volumetric cutting in noise reduction, which can improve the quality of reconstruction.

\section{Conclusion}
We have demonstrated that our point-based representation, Gaussian surfels, can be efficiently optimized to achieve a good balance between high-quality surface reconstruction and computational cost.

\paragraph{Limitations} Even with monocular norm priors, our method cannot guarantee accurate reconstruction results at areas with strong specular reflections. In the future, we will investigate how to store and optimize features at Gaussian surfels to encode their view-dependent appearance. Taking the features and view direction as inputs, we can train a neural network decoder to give us additional capabilities to handle specular reflections than spherical harmonics. Moreover, we have also observed that, for surfaces with very weak textures, our reconstructed surfaces may exhibit a global shift compared to ground-truth surfaces. It is possible to mitigate this issue by incorporating information from depth sensors or more shape priors.

\begin{acks}
We thank the anonymous reviewers for their professional and constructive comments. Weiwei Xu is partially supported by NSFC grant No.~61732016. Jiamin Xu is partially supported by NSFC grant No.~62302134 and ZJNSF grant No.~LQ24F020031. This paper is supported by Information Technology Center and State Key Lab of CAD\&CG, Zhejiang University.
\end{acks}

\bibliographystyle{ACM-Reference-Format}
\bibliography{reference}

\end{document}